# Towards Quantum Federated Learning

Chao Ren, *Member*, *IEEE*, Rudai Yan, *Member*, *IEEE*, Huihui Zhu, Han Yu, *Senior Member*, *IEEE*, Minrui Xu, Yuan Shen, Yan Xu, *Senior Member*, *IEEE*, Ming Xiao, *Senior Member*, *IEEE*, Zhao Yang Dong, *Fellow*, *IEEE*, Mikael Skoglund, *Fellow*, *IEEE*, Dusit Niyato, *Fellow*, *IEEE*, and Leong Chuan Kwek

***Abstract*— Quantum Federated Learning (QFL) is an emerging interdisciplinary field that merges the principles of Quantum Computing (QC) and Federated Learning (FL), with the goal of leveraging quantum technologies to enhance privacy, security, and efficiency in the learning process. Currently, there is no comprehensive survey for this interdisciplinary field. This review offers a thorough, holistic examination of QFL. We aim to provide a comprehensive understanding of the principles, techniques, and emerging applications of QFL. We discuss the current state of research in this rapidly evolving field, identify challenges and opportunities associated with integrating these technologies, and outline future directions and open research questions. We propose a unique taxonomy of QFL techniques, categorized according to their characteristics and the quantum techniques employed. As the field of QFL continues to progress, we can anticipate further breakthroughs and applications across various industries, driving innovation and addressing challenges related to data privacy, security, and resource optimization. This review serves as a first-of-its-kind comprehensive guide for researchers and practitioners interested in understanding and advancing the field of QFL.**

## I. Introduction

### A. Backgrounds

QUANTUM Federated Learning (QFL) is a collaborative distributed learning framework that leverages characteristics of quantum mechanics to address challenges such as high computational cost and security of transmitted messages in the federated learning (FL) process. It leverages the characteristics of quantum computing (QC), such as quantum parallelism, superposition, entanglement and so forth, to solve those challenging issues faced by classical machine learning (ML) methods and provide potential improvements in computational speed, security and privacy.

The rapid development of QC and FL technologies have given rise to two fields that hold the potential to significantly reshape the landscape of computation and ML. QC is a computational paradigm that leverages the principles of quantum mechanics to perform complex computations more efficiently than classical computers [1]. QC has attracted much interest in the 'Noisy Intermediate-Scale Quantum-computers' (NISQ) era [2], [3], [4], due to quantum advantages in solving specific computationally complex problems using various hardware architectures, including trapped ion systems [5], [6], superconducting systems [7], [8], measurement-based QC [9], [10], and Gaussian boson sampling on photonic platforms [11]. Researchers have identified several quantum algorithms that outperform the best classical algorithms, including Shor's algorithm for large prime number factorization [12] and Grover's algorithm for unstructured search [13]. By exploiting quantum phenomena, these QC algorithms offer the potential to reduce computational complexity and enhance security and privacy mechanisms to solve problems in ML [14], [15], science [16], [17], and other areas [18], [19]. Meanwhile, FL has emerged as a collaborative learning approach that addresses user privacy and data confidentiality concerns. It enables the training of ML models across multiple devices or organizations, each with its own local dataset, while avoiding the exchange of raw data [20]. FL has found applications in various domains, including healthcare [21], finance [22], and smart cities [23], [24], where data privacy and security are of paramount importance.

By integrating QC and FL, QFL can unlock new opportunities and challenges for ML and various application domains, such as healthcare and medical diagnosis, finance and risk assessment, telecommunications and networking, smart cities and transportation, environmental monitoring and climate modeling, *etc*. By combining the computational advantages of QC with the privacy-preserving capability of FL, QFL has the potential to transform the way ML models are developed, trained, and deployed, making it possible to build more secure, efficient, and widely applicable solutions that can tackle real-

C. Ren is currently a Wallenberg-NTU Presidential Postdoctoral Fellow at the School of Electrical Engineering and Computer Science, KTH Royal Institute of Technology, Sweden; and the School of Electrical and Electronic Engineering, Nanyang Technological University, Singapore.

R. Yan, Y. Shen, and H. Zhu are with the School of Electrical and Electronic Engineering, Nanyang Technological University, Singapore.

H. Yu is a Nanyang Assistant Professor at the College of Computing and Data Science, Nanyang Technological University, Singapore. (*Corresponding Author*, email: han.yu@ntu.edu.sg)

M. Xu is with the College of Computing and Data Science, Nanyang Technological University, Singapore.

Y. Xu is an Associate Professor at the School of Electrical and Electronic Engineering, Nanyang Technological University, Singapore.

M. Xiao is a Full Professor at the School of Electrical Engineering and Computer Science, KTH Royal Institute of Technology, Sweden.

M. Skoglund is Head of the Division of Information Science and Engineering at the KTH Royal Institute of Technology, Sweden.

Z.Y. Dong is the Head of Department and Chair Professor of Electrical Engineering, City University of Hong Kong, China.

D. Niyato is a President's Chair Professor at the College of Computing and Data Science, Nanyang Technological University, Singapore.

L.C. Kwek is currently a Principal Investigator at the Center for Quantum Technologies, National University of Singapore, and a co-Director of the Quantum Science and Engineering Centre (QSEC) at the Nanyang Technological University, Singapore.

TABLE I
SUMMARY OF RELATED REVIEWS VERSUS OUR QFL REVIEW

| Year | References | Key Contributions | QFL Algorithms | QFL Application | QFL Taxonomy | QFL Challenge | QFL Future Directions |
|---|---|---|---|---|---|---|---|
| 2021 | Chen *et al.*, [25] | Provided a framework and process for federated training on hybrid quantum-classical classifiers | √ | - | - | - | - |
| 2022 | Larasati *et al.*, [26] | Provided a basic overview and the potential of QFL, and discussed the unique characteristics in relation to QFL development | - | √ | - | √ | - |
| | Kwak *et al.*, [27] | Compared several model structures for QDDL. The authors explore the possibilities and limitations of QDDL for various applications | √ | √ | - | √ | - |
| 2023 | Qiao *et al.*, [28] | Providing an in-depth exploration of the various facets of FL paradigms, ultimately leading to a thorough understanding of QFL | √ | √ | - | √ | √ |
| | ours | Proposed a detailed taxonomy for QFL, identify the challenge and open issues of QFL, and give several valuable future research directions | √ | √ | √ | √ | √ |

world problems while protecting sensitive data and encouraging collaboration across different organizations.

### *B. Motivations for QFL*

The motivation for QFL stems from the desire to address the limitations of classical FL and harness the potential of QC to improve it. It has the potential to offer several advantages over classical FL approaches. The key motivating factors for developing QFL include:

**A1)** *Enhancing Privacy and Security*: Although FL is designed to keep raw data on local devices, sharing model updates between devices can still expose sensitive information. Classical encryption techniques can provide a certain level of protection, but they may not be robust enough against increasingly sophisticated attacks. QFL can enhance privacy by using quantum-secured techniques, like quantum key distribution (QKD), which is more secure than classical encryption methods and to provide a higher level of privacy and security guarantee against eavesdropping and hacking.

**A2)** *Improving Computational Efficiency*: Training complex FL models can be time-consuming and resource-intensive, especially when dealing with large datasets and models. Classical FL algorithms can be limited by the computational capabilities of local devices, leading to slow convergence or suboptimal model performance. QFL can potentially improve computational efficiency by leveraging the unique capabilities of quantum computers, such as quantum parallelism and entanglement, to speed up the training FL process, which can be particularly beneficial in scenarios where fast decision-making is crucial.

**A3)** *Reducing Communication Overhead*: In classical FL, the process of transmitting model updates between devices can be bandwidth-intensive and slow down the FL process, particularly when dealing with large models and frequent updates. This can lead to high communication costs and slow down the overall FL process. QFL could help reduce the communication overhead by compressing model updates using quantum techniques, such as quantum data encoding, which enable more efficient model updates and faster convergence. and transmit information more compactly than classical FL methods.

**A4)** *Improving Model Performance*: Classical FL methods might struggle to achieve optimal model performance in some cases, particularly when dealing with non-convex optimization problems, which are common in deep learning (DL). QFL has the potential to offer more efficient optimization algorithms by exploiting quantum parallelism and quantum entanglement, which can result in better model performance and generalization in federated settings.

**A5)** *Enhancing Scalability*: With the rapid growth of data and model complexity, classical FL methods may struggle to scale efficiently. QFL seeks to harness the inherent parallelism of quantum computers to process massive datasets and train complex models across numerous nodes, enabling more scalable FL.

**A6)** *Enabling New Applications*: QC enables the development of novel ML algorithms and applications that were previously considered computationally infeasible with classical computing. QFL can bring these new algorithms to distributed settings, opening up new possibilities in various fields such as healthcare, finance, and transportation, and more.

### *C. Related Works and Contributions*

In this article, we provide an overview of research works related to QFL. Due to the growing research interest surrounding QFL, there only have been three recent short reviews on associated subjects lately. Table I shows a comparison between these reviews and our work. The short review in [25] discusses the concept of federated quantum machine learning (QML), which involves training QML models in a distributed manner across multiple quantum computers. The authors show a framework for federated training on hybrid quantum-classical classifiers, where they consider the quantum neural network (QNN) coupled with classical pre-trained convolutional model. They demonstrate that this approach can achieve almost the same level of trained model accuracies and yet significantly faster distributed training. The authors also address the challenges associated with federated QML, including privacy concerns and the need for secure infrastructure. They show that their proposed framework can help build secure QML infrastructure and better utilize available NISQ devices. The paper in [26] provides a basic overview and the potential of QFL. The authors discuss the unique characteristics of QC and how they differ from classical computation, which poses even more challenges for a federated setting. They also explore the limitations of QC in relation to QFL development, such as the difficulty in implementing error correction and scalability issues. Additionally, possible approaches to deploy QFL are explored, including hybrid classical-quantum approaches and distributed QC. The authors also present remarks and challenges of QFL, such as privacy concerns and the need for standardized protocols. The review in [27] discusses the emerging field of quantum distributed DL (QDDL) and compares several model structures for QDDL. The authors explore the possibilities and limitations of QDDL for various application scenarios, and highlight the benefits of using QDDL, such as improved data security and reduced computational overload, compared to traditional ML methods. The latest survey in [28] encompasses the entire spectrum of FL paradigms, from fundamental FL concepts to the cutting-edge developments in QFL, but still no reasonable categorization has been proposed.

While the above-mentioned reviews provide the good overview of some aspects of QFL, they are brief and often lack comprehensiveness. Moreover, they are limited in the consideration of QML aspects. Only [27] takes into account the issue of quantum-secured mechanisms (QSMs), but they all neglect the issue of quantum optimization algorithms (QOAs). While these reviews have focused primarily on QML for QFL, it is essential to consider QSM and QOA as well to gain a holistic understanding of the field.

Different from existing reviews, this article comprehensively analyzes the three key aspects of QFL, including QNNs within QML, QSMs and QOAs. In addition, we analysis the current state of QFL, its underlying concepts, the challenges it faces, give a detailed literature review for recent advances of QFL and summarize them from several aspects. Furthermore, we propose a taxonomy of the QFL literature from multiple aspects. Finally, we discuss the challenges and open issues for current QFL, and then identify future research directions and open problems in QFL, emphasizing the need for developing QFL to fully realize its potential. The key contributions of our review are as follows.

1) We provide a concise overview of FL and its classification, quantum algorithms and complexity in terms of QNNs within QML, QSMs, and QOAs, along with an in-depth analysis of the primary motivations for QFL in the current FL settings.
2) We identify QC techniques to tackle key FL challenges and propose a detailed taxonomy of QFL in terms of different kinds of FL perspective to guide the review of QFL literature. Based on this perspective, we propose a hierarchical taxonomy to showcase existing QFL works, emphasizing the challenges they encounter, their core ideas, and key contributions.
3) We discuss the evaluation metrics and potential platforms in the current QFL literature for benchmarking purposes in order to give the guide for future QFL algorithms. Besides, we summarize the frequently adopted public datasets and corresponding application of current QFL literature.
4) We highlight challenges and open issues of QFL, then we anticipate promising future research directions, focusing on quantum hardware limitations, noise and error mitigation, model and data heterogeneity, interoperability between quantum and classical FL, standardization and benchmarking, and ethics and legal considerations towards developing more robust, efficient, effective, interpretable, fair, personalized, secured, inductive and comprehensive QFL systems.

## II. Terminology

The general framework of QFL can be conceptualized as a distributed ML approach where multiple quantum nodes (each possessing QC capabilities, e.g., QNNs within QML, QSMs, QOAs) collaboratively learn a model while keeping the data localized safely. This framework inherently respects data privacy, leverages QC advantages, and allows for the exploitation of quantum properties such as superposition and entanglement to enhance model training and inference processes, potentially surpassing classical FL capabilities. Mathematically, the QFL process is similar with classical FL and can be described as follows:

- Each quantum node $i$ holds a local dataset $D_i$ and initializes a local model $M_i$ by QNNs.
- These local models are trained on their respective datasets to minimize a local loss function $L(M_i, D_i)$ by QOAs.
- Periodically, the secure aggregation protocol by QOAs or QSMs is employed to compute a global model $M_{global}$ by aggregating the parameters of the local models without exposing individual data points. This can be represented as $M_{global} = f(M_1, M_2, \dots, M_n)$, where $f$ represents the aggregation function and $n$ is the number of nodes.
- The global model is then shared back with each node for further local updates by QSMs.

This section aims to effectively connect the fundamental concepts of FL and QC with the specific techniques relevant to QFL, thus ensuring that readers are well-equipped with the necessary background as they proceed to Section III and beyond. Section II.A introduces the four categorizations of FL and their sub-categories. The purpose is to provide a comprehensive overview of FL, highlighting the diversity and scope of methods within this field. Section II.B presents fundamental concepts of QC to acquaint readers with the technological underpinnings that facilitate QFL. This section is crucial for understanding how quantum advancements can enhance FL. In Section II.C, we then transition to discuss three specific QC techniques, including QNNs within QML, QSMs, and QOAs. Each of these is explored in the context of their applicability and potential in enhancing FL through quantum approaches. The discussion on QNNs within QML, QSMs, and QOAs includes an exploration of why these particular quantum techniques are suitable for QFL, supplemented with references to key original research works.

### *A. Basics of FL*

FL is a distributed ML approach that enables multiple parties to collaboratively train a shared model while keeping their data locally, thereby preserving data privacy. It can be classified based on various factors, such as the data distribution, network architecture, privacy mechanism, and model aggregation techniques [29]. Here are some common classifications of FL:

**F1)** *Data Distribution*: Based on data distribution, FL can be classified into horizontal FL (HFL), vertical FL (VFL), and federated transfer learning (FTL). HFL is used when clients have the same feature space but different samples (i.e., IID data) [30]. In this scenario, the clients collaborate to train a model without sharing raw data. HFL is particularly useful when multiple organizations or individuals are needed to train a model collaboratively without sharing sensitive information. VFL is used when clients have different feature spaces but share the same sample set. Clients jointly train a model without sharing raw feature data. VFL enables the combination of complementary feature sets from different clients to train a more accurate and robust model [31]. FTL combines FL and transfer learning, where knowledge learned in one domain is transferred to another domain while preserving data privacy [32]. It enables clients to take advantage of pre-trained models or knowledge from other domains to improve model training efficiency and performance.

**F2)** *Network Architecture*: Based on network architecture, FL can be classified into centralized FL (CFL) and decentralized FL (DFL). For CFL, clients communicate with a central server for model aggregation. Clients train local models using their data and send the model updates to the server [20], [33]. The server aggregates these updates and returns the updated global model to the clients. For DFL, clients directly communicate with each other, without the need for a central server. Clients exchange model updates in a peer-to-peer fashion, and model aggregation is performed in a distributed manner [34]. DFL can enhance privacy and reduce the risk of a single point of failure.

**F3)** *Privacy Mechanism*: Based on privacy mechanism, FL can be classified into differential privacy (DP)-based FL (DPFL) and encrypted-based FL (EFL). DPFL incorporates DP techniques to ensure that individual data points cannot be inferred from the aggregated model updates [35]. Noise is added to the model updates during the FL training process, providing strong privacy guarantees while maintaining model utility. EFL uses encryption techniques, such as homomorphic encryption (HE) or secure multi-party computation (SMPC), to protect the privacy of model updates during the FL training process [36]. Clients submit encrypted model updates that can be aggregated without being decrypted, ensuring data privacy without compromising the quality of the global model.

**F4)** *Model Aggregation*: There are several widely adopted FL model aggregation algorithms. Federated averaging (FedAvg) calculates the weighted average of local model updates to create a global model [37]. Each client's local model update is weighted by the number of training data samples, ensuring that the global model reflects the contributions of all clients. Federated stochastic gradient descent (FedSGD) updates the global model using stochastic gradient descent (SGD) based on the local gradients from each client [38]. The server aggregates the gradients from clients and applies the updates to the global model. While FedSGD transmits smaller gradients per round, it requires more frequent communication. FedAvg reduces communication frequency by performing multiple local updates before aggregation, making it potentially more efficient overall. Other advanced aggregation techniques can involve using more sophisticated aggregation methods, such as geometric median, trimmed mean, or robust aggregation algorithms, to improve the quality of the global model and mitigate the effects of noisy or malicious clients [39].

### *B. Basics of Quantum Computing*

QC exploits phenomena of quantum nature, such as superposition and entanglement, to provide beyond-classical computational capabilities. It relies on the principles of quantum mechanics to perform complex computations more efficiently than classical computers for specific problem domains [40]. This section provides a general narrative on basic QC concepts.

➢ *Qubits*: The fundamental building block of QC is the quantum bit, or qubit, which, unlike classical bits, can represent not only 0 and 1, but also a combination of both states [41]. Mathematically, a qubit can be described as a linear combination of its basis states $|0\rangle$ and $|1\rangle$:

$$|\psi\rangle = \alpha|0\rangle + \beta|1\rangle \quad (1)$$

where $\alpha$ and $\beta$ are complex numbers satisfying $|\alpha|^2 + |\beta|^2 = 1$ . This unique property allows quantum computers to process a vast amount of information simultaneously by encoding multiple possibilities in a single qubit. With multiple qubits, the dimension of the total Hilbert space accessible become exponentially large, thus enabling them to solve problems that are intractable for classical computers [42].

➢ *Quantum Superposition*: As shown in Eq. (1), the general state of a qubit can be formed as the linear combinations of states. Such characteristic is often called superposition. Quantum superposition is a fundamental concept in quantum mechanics [1], which states that a quantum system may be in one of many configurations (arrangements of particles or field) and the most general state is a combination of all of these possibilities, where the amount in each configuration is specified by a complex number. This phenomenon is derived from the wave-like nature of quantum particles, such as electrons or photons, which allows them to occupy different states of position, energy, or other properties at the same time. Mathematically, the state of a quantum system is represented by a vector in a complex Hilbert space, and the linear structure of the Hilbert space implies that any linear combination of these state vectors is also a valid state of the system. Superposition is crucial for understanding the quantum systems and is a key resource underlying many quantum phenomena, such as quantum parallelism and quantum entanglement.

➢ *Quantum Entanglement*: Quantum entanglement is a unique phenomenon in which the states of two or more quantum particles become intertwined, such that the state of one particle cannot be described independently of that of the other(s) [43]. This property arises due to the superposition principle and has profound implications for quantum information technologies. Entangled qubits can be created through operations like a Hadamard (H) gate followed by a controlled-NOT (CNOT) gate and can be utilized to perform complex, correlated operations on multiple qubits simultaneously. Quantum entanglement enables more efficient computation, as well as novel protocols for secure communication and distributed computing [44].

➢ *Quantum Measurement*: Quantum measurement is a key concept in quantum mechanics that describes the process of observing or measuring a quantum system [45]. A closed quantum system evolves according to unitary evolution. Upon measurement, the quantum system is no longer closed and will collapse into a state with probabilities determined by the squared magnitudes of the coefficients associated with each measurement outcome. This collapse is inherently probabilistic, and the outcome cannot be predicted with certainty.

➢ *Quantum Gates and Circuits*: Quantum gates are the fundamental operations used to manipulate the states of qubits in a controlled manner, utilizing the principles of superposition, entanglement and measurement [46]. Any quantum operation (unitary transformation) can be implemented to arbitrary precision by a sequence of gates. Some common quantum gates include the Pauli-X, Y and Z, Hadamard, phase S and CNOT gates. These gates are referred to as the Clifford group. However, any quantum circuit consists of only Clifford group can be efficiently simulated on a classical computer. Therefore, in order to show quantum speedup, it is essential to include non-Clifford gates such as Toffoli (CCNOT) gate. Notably, quantum gates are reversible, meaning that they can transform a quantum state back to its original state, and the inverse of a quantum gate can be easily computed [47].

## C. Quantum Algorithms and Complexity

Quantum algorithms exploit the unique properties of quantum phenomena, such as superposition and entanglement, to achieve significant speedups over classical algorithms for complex problems. Quantum complexity theory investigates the computational resources required by quantum algorithms, often expressed in terms of the number of qubits and the number of quantum gate operations. Complexity classes, such as BQP (bounded-error quantum polynomial time), represent problems that can be efficiently solved by quantum computers. BQP is believed to be larger than classical complexity classes, implying that some problems are more efficiently solvable using quantum computers than classical ones [48].

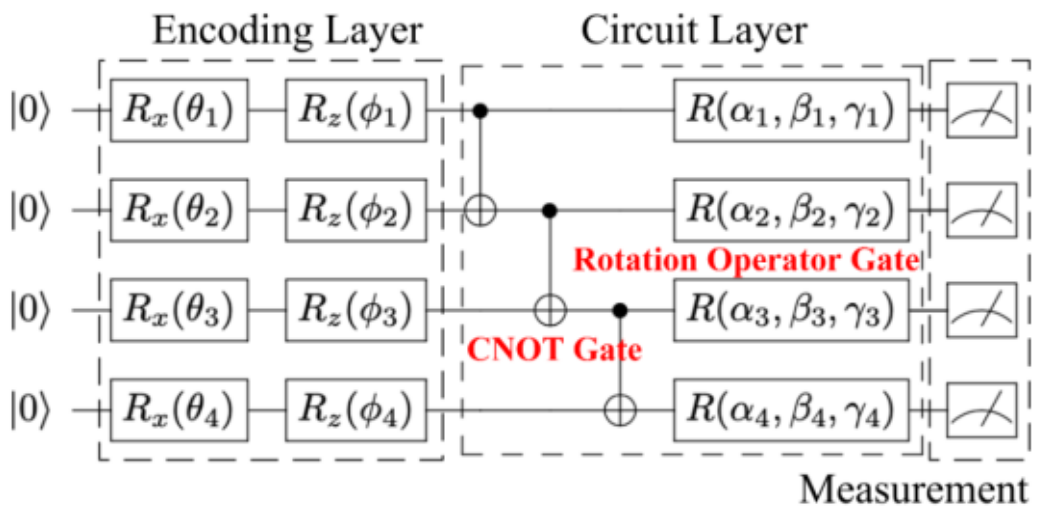

Fig. 1. The basic structure of VQC.

In this section, we will describe quantum algorithms and complexity from three perspectives, including QNNs within QML, QSMs, and QOAs.

### Q1) *Quantum Neural Network (QNN) within QML*

Due to the limited availability of large-scale quantum computers and the significant advances in classical ML, hybrid quantum-classical ML that integrates quantum circuits with classical neural network (NN) has attracted much attention. For example, Ref. [49] introduce the quantum deep NN (QDNN) which is a composition of multiple QNN layers. QDNN can keep the advantages of the classical NN such as the non-linear activation, the multi-layer structure, and the efficient backpropagation training algorithm. The inputs and the outputs of the QDNN are both classical which makes the QDNN more practical. Ref. [50] proposes a hybrid quantum-classical NN architecture where each neuron is a variational quantum circuit. Such combination leads to the development of various QNNs, which may have promising performance in some tasks. This section explicitly states that its primary focus is on QNNs and their relevance to QFL, rather than on QML as a whole. We provide a more detailed introduction explaining the potential QNNs fitting into the broader of QFL.

As for the implementation, variational quantum circuit (VQC) [51] is a type of quantum circuit with adjustable parameters that is commonly used to formulate a QNN in hybrid quantum-classical ML, leveraging the principles of QC to perform learning tasks (including classification and regression problem) [52]. The parameters can be trained by a classical optimization algorithm/solver to find the optimal set of parameters that minimizes a certain cost or objective function [53]. Fig. 1 shows a typical structure of VQC which consists of two kinds of quantum gates, i.e., CNOT gate and rotation operator gate. The rotation operator gates $R_x(\theta)$, $R_y(\theta)$, and $R_z(\theta)$ are the rotation matrices in three Cartesian axes, $\theta$ is the rotation angle w.r.t a certain axis. This VQC includes two layers, encoding layer and circuit layer. The encoding layer encodes the classical data as the quantum states for further processing in circuit layer. Note that a VQC is not only limited to the form in Fig. 1 and can have various structures based on the designed and targeted

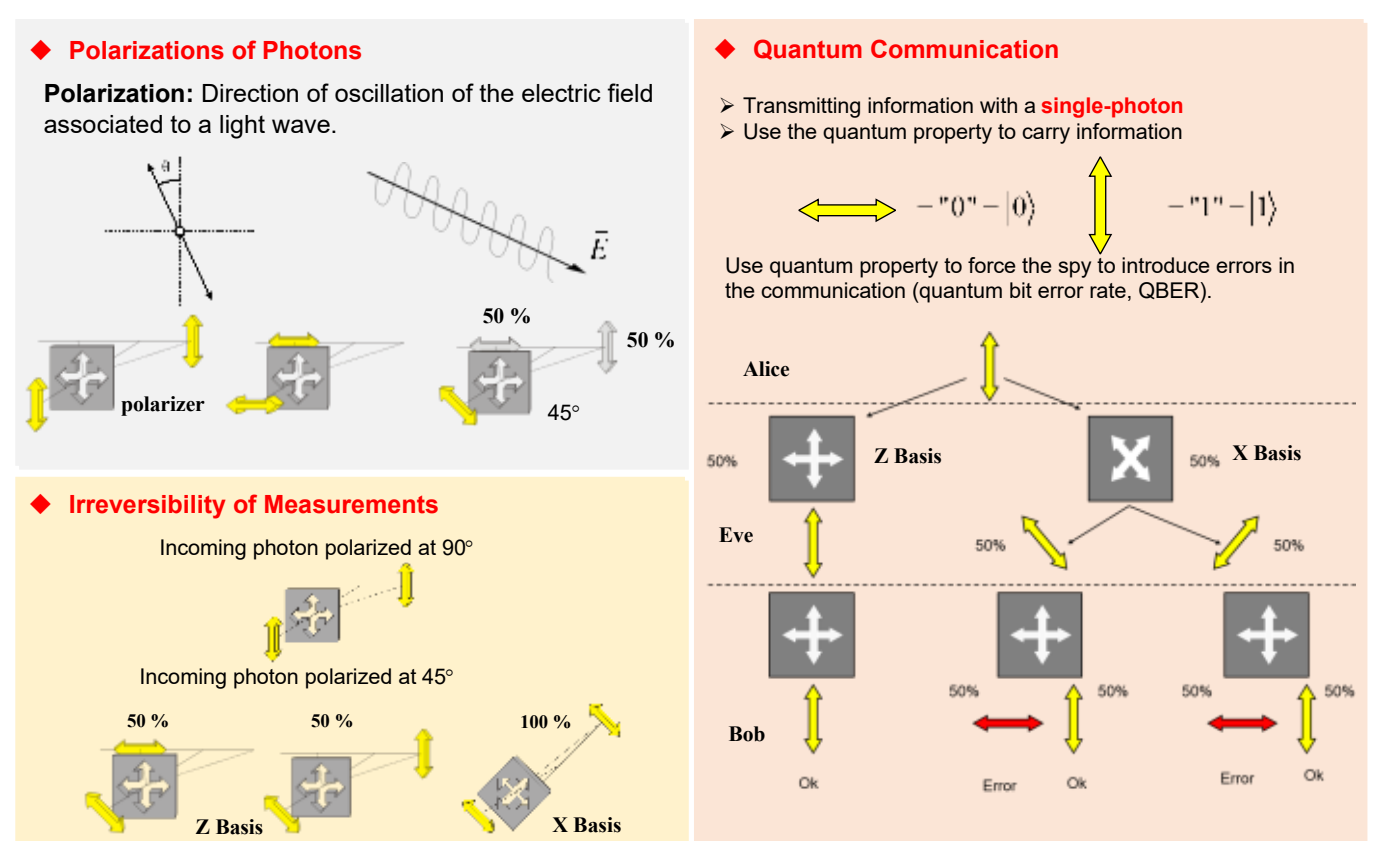

Fig. 2. Principles of quantum communications with basic ingredients of BB84 protocol.

problem. Several representative QNNs that can be realized by VQCs are reviewed in the following.

➢ *Quantum Convolutional Neural Networks (QCNN)* [54]: QCNN is an area of research that explores the potential of QC to accelerate the training and inference of neural network (NN). Ref. [9] proposes a quantum version of the convolutional NN (CNN), which is a widely used architecture in classical ML. QCNN can achieve better performance than classical CNN on certain image recognition tasks.

➢ *Quantum Long Short-Term Memory (QLSTM)* [55]: Ref. [55] extends the classical long short-term memory (LSTM) into the quantum realm by replacing the classical NN in the LSTM cells with VQC, which would play the roles of both feature extraction and data compression.

➢ *Quantum Generative Adversarial Network (QGAN)* [56]: QGAN is an emerging area of research that aims to apply the principles of QC to the field of generative modeling. Ref. [56], [57] introduce QGAN, where the data consist of quantum states or classical data, and the generator and discriminator are equipped with quantum information processors.

➢ *Quantum Transfer Learning* [58]: Ref. [58] extends the concept of transfer learning, widely applied in modern ML algorithms, to the emerging context of hybrid NN composed of classical and quantum elements. This paper proposes different implementations of hybrid transfer learning, but we focus mainly on the paradigm in which a pre-trained classical NN is modified and augmented by a final VQC.

**Q2) *Quantum-Secured Mechanisms (QSM)***

QSM approaches rely on quantum communication to enhance the security in data transmission. Fig. 2 briefly explains the principle of quantum communication in the context of photon, which is the main information carrier in quantum communication. The security promised by quantum communication is ensured by the irreversibility of measurements. Conventionally, the classical bit is encoded in the direction of polarization. As Fig. 2 shows, measurements in X-basis performed by the eavesdropper Eve will change the original direction of polarization to two possible directions with the probability of 50%. Therefore, any measurement performed by an eavesdropper will result in a higher quantum bit error rate. quantum key distribution (QKD) and quantum secure direct communication (QSDC) are both secure communication protocols that leverage the principles of quantum communication mentioned above to provide security against eavesdropping. While QKD focuses on the exchange of cryptographic keys, QSDC allows for the direct transmission of secret messages.

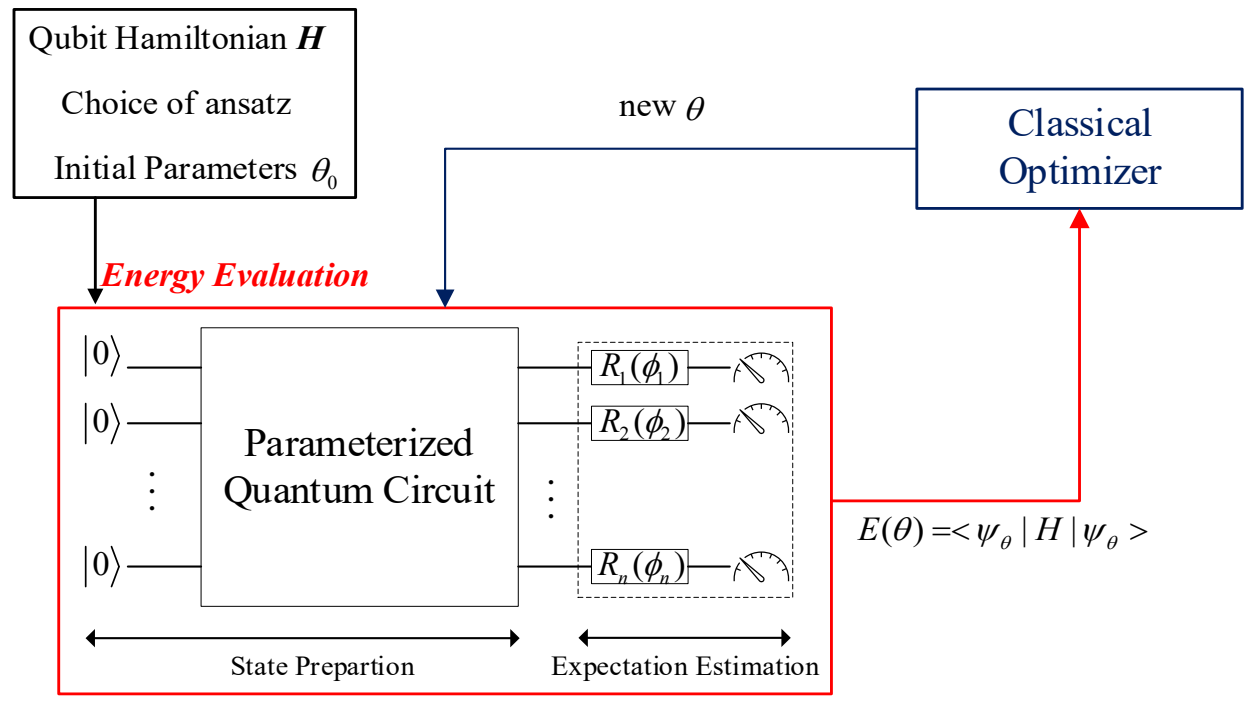

Fig. 3. Basic structure of VQE.

➢ *Quantum Key Distribution (QKD)* [59]: QKD is designed to exchange cryptographic keys between two parties. It enables two communication parties (usually denoted as Alice and Bob) to generate a common secret key in the presence of an eavesdropper (Eve), which is based on transmitting non-orthogonal quantum states. To communicate classical messages and conduct post-processing after transmitting and measuring these quantum states to produce a key that is secure. To prevent the man-in-the-middle attack, the communicating parties need to authenticate these classical messages in advance. In QKD, the communicating parties exchange a random sequence of qubits encoded in different bases. After the exchange, they publicly announce the bases they used and compare them. Whenever their bases match, the corresponding bits become part of the shared secret key [60]. Due to the fundamental principles of quantum mechanics, such as the no-cloning theorem and Heisenberg's uncertainty principle, any attempt by an Eve to measure the exchanged qubits will introduce detectable errors, revealing her presence. There are different types of QKD protocols, such as prepare-and-measure protocols (e.g., BB84 [59]) and entanglement-based protocols (e.g., Ekert91 [61]). These protocols can be implemented using various physical systems, such as photons, to encode and transmit cross-out information.

➢ *Quantum Secure Direct Communication (QSDC)* [62]: QSDC allows direct transmission of secret messages without the need to establish a shared key beforehand [63], [64]. Actually, the first QSDC protocol was proposed in [62], it can transmit deterministic key, which also transmit the secret message. Unlike QKD, where a secret key is exchanged and then used for encrypting and decrypting messages, QSDC enables secure message transmission in a single step. In 2003, the standard of QSDC was proposed [65]. Recently, QSDC was developed in both theory and experiment. In theory, the one-step QSDC protocol was proposed [66]. It only requires

TABLE II

TYPES OF QC AND FL ALGORITHMS, PURPOSE, AND ADVANTAGES OF DIFFERENT QFL CATEGORIES

| **Methods**<br>**√ : suitable; ○: suitable but with none or only less work; -: not suitable** | | | **Purpose and Advantages** | | | | | | **QC** | | | **FL** | | | |
|---|---|---|---|---|---|---|---|---|---|---|---|---|---|---|---|
| | | | A1 | A2 | A3 | A4 | A5 | A6 | Q1 | Q2 | Q3 | F1 | F2 | F3 | F4 |
| **EffQFL** | data processing-based | | √ | √ | √ | √ | √ | √ | √ | - | ○ | √ | ○ | - | - |
| | model optimization -based | global | - | √ | √ | √ | √ | √ | ○ | - | √ | √ | ○ | - | √ |
| | | local | - | √ | - | √ | √ | √ | √ | - | √ | √ | ○ | - | - |
| | client selection-based | | - | √ | √ | - | √ | √ | ○ | - | ○ | ○ | ○ | - | √ |
| **SecQFL** | data privacy-based | | √ | - | - | - | √ | √ | ○ | ○ | - | ○ | ○ | ○ | - |
| | model security-based | | √ | - | - | - | √ | √ | - | √ | - | ○ | ○ | √ | - |
| | robustness-based | | √ | - | - | √ | √ | √ | ○ | √ | - | √ | ○ | √ | ○ |

A1: Enhanced Privacy and Security
A2: Improved Computational Efficiency
A3: Reduced Communication Overhead
A4: Better Model Performance
A5: Scalability
A6: Enabling New Applications
Q1: QNN
Q2: QSM
Q3: QOA
F1: Data Distribution
F2: Network Architecture
F3: Privacy Mechanism
F4: Model Aggregation

transmitting the entanglement for one round and the existing QSDC protocols all need to transmit two rounds. The one-step device-independent QSDC protocol [67] and measurement-device-independent QSDC protocol were also proposed [68]. In experiment, the QSDC can reach 100 km in fiber [69]. The security of QSDC is based on the quantum mechanics principles, such as quantum entanglement and the no-cloning theorem. QSDC protocols have been proposed for secure voting, quantum secret sharing, and quantum private comparison.

➢ *Quantum Error Correction (QEC)* [70]: QEC is a set of techniques used to protect quantum information from errors caused by the noise and decoherence in quantum systems. Unlike classical information, which can be copied and read without being disturbed, quantum information is inherently fragile and susceptible to errors, and cannot be copied with unit fidelity according to the no-cloning theorem. QEC is based on the principle of redundancy, where multiple copies of a quantum state are stored and processed in a way that allows errors to be detected and corrected [71], [72]. The basic idea is to encode the logical qubit onto a larger number of physical qubits, and to use quantum stabilizers to detect errors that occur during the computation, then correct the error with single qubit operations. The effectiveness of QEC depends on the quality of the quantum codes used, as well as the quality of the quantum operations used for error detection and correction.

**Q3)** ***Quantum Optimization Algorithm (QOA)***

QOA aims to use quantum methods (i.e., the quantum nature of multiple photons, including quantum superposition, interference and entanglement) to tackle challenging optimization problems that possess comparable complexity (NP-hard) to notoriously tough problems that have resisted efficient solutions for decades. Two typical QOAs can be applied for ML tasks are reviewed in the following.

➢ *Variational Quantum Eigensolver (VQE)* [73]: Fig. 3 illustrates the idea of VQE. The basic idea of the VQE algorithm is to prepare a trial wavefunction on a quantum computer using a set of parameterized quantum gates. These gates are chosen to be simple and efficient, such as rotations around the X, Y, and Z axes, which can be implemented using Pauli gates. The parameters of these gates are then optimized classically in order to minimize the energy expectation value. VQE algorithm is a powerful tool for studying molecular properties because it can be used to find the ground state energy of molecules that are too large to simulate using classical computers. VQE can be adapted to solve optimization problems by constructing a Hamiltonian that represents the cost function of the problem. By finding the ground state energy of this Hamiltonian, VQE essentially finds the optimal solution to the optimization problem [74]. VQE works by first encoding the problem of finding the ground state energy of a molecule into a quantum circuit. This circuit is then run on a quantum computer, which produces an output state that approximates the ground state of the molecule. The output state is then measured, and the resulting measurements are used to compute an estimate of the energy of the molecule. This energy estimate is then fed back into the circuit, which is updated to produce a new output state that is closer to the true ground state energy.

➢ *Quantum Approximate Optimization Algorithm (QAOA)* [75]: QAOA is designed to solve combinatorial optimization problems that involve finding the optimal solution from a finite set of possible solutions. The Hamiltonian, as a mathematical representation of the energy of a quantum system, is constructed using a set of objective and constraint functions that describe the problem. The algorithm then applies a sequence of quantum gates to the quantum state of the system, which evolves the system towards the ground state of the Hamiltonian [76]. The ground state of the Hamiltonian corresponds to the optimal solution of the original combinatorial optimization problem. QAOA combines classical and quantum computations and the classical part of QAOA involves optimizing the parameters of the quantum gates in order to minimize the energy of the system.

## III. PROPOSED QFL TAXONOMY

In our systematic and comprehensive review of current and potential QFL approaches, we propose a taxonomy that classifies QFL into two categories: 1) **Efficient QFL** (EffQFL) and 2) **Secured QFL** (SecQFL), as shown in Fig. 4. This taxonomy is based on the characteristics and quantum algorithms involved in each category. The purpose of EffQFL and SecQFL is to address two crucial aspects of QFL respectively that are critical for the successful application and adoption of QC in FL tasks. These two aspects focus on different, yet complementary, objectives. In Table II, we summarize the proposed taxonomy together with QFL purpose and advantages. Below is the brief introduction to EffQFL and SecQFL approaches.

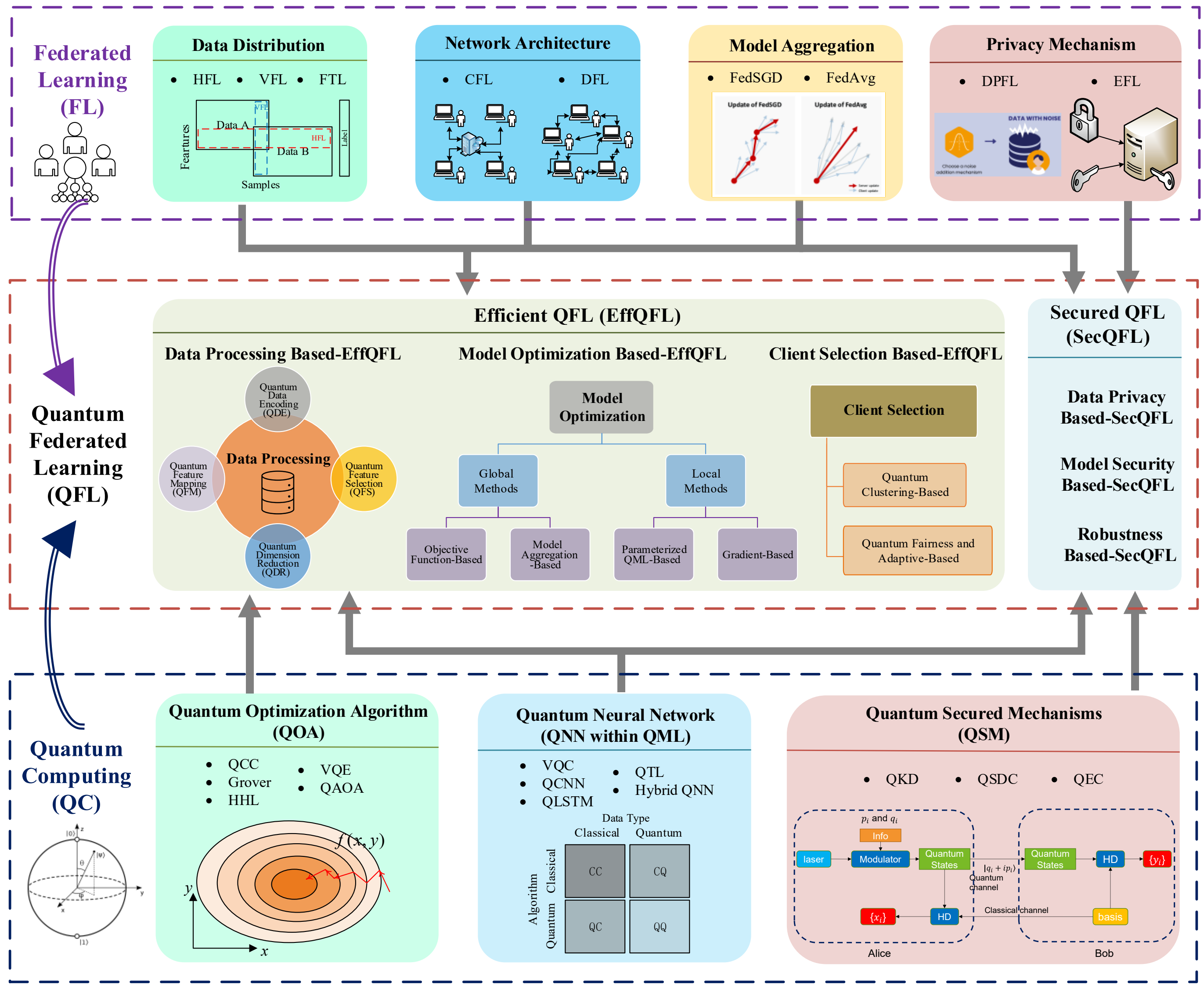

Fig. 4. The proposed QFL taxonomy.

## *A. EffQFL*

EffQFL is to maximize the potential of QC for learning features from data more efficiently and effectively compared to classical FL. The goal is to achieve faster training, improved scalability, and better performance for FL by utilizing QOA and QNN. EffQFL can mainly address the following issues:

➢ Scalability: Classical FL methods can struggle with scalability when dealing with large datasets or high-dimensional data. EffQFL aims to develop QOA and QNN that can efficiently handle large-scale problems and improve computational complexity.

➢ Speed: Training complex FL models can be time-consuming. EffQFL focuses on leveraging quantum properties like superposition, entanglement, and quantum interference to speed up the learning process.

➢ Expressiveness: Finding suitable feature representations is crucial for FL performance. EffQFL investigates quantum feature maps (QFM) and quantum circuits that can generate complex and high-dimensional feature representations, enabling better model performance.

➢ Resource Efficiency: QC resources are currently limited, and designing efficient QOA and QNN is essential. EffQFL develops resource-efficient methods that can be practically implemented on existing quantum hardware.

EffQFL for this category are classified into data processing-based, model optimization-based, and client selection-based approaches. Data processing-based EffQFL focuses on investigating quantum version of feature engineering. Model optimization-based EffQFL consists of the global part (including objective function-based and model aggregation-based approaches) and the local part (including parameterized QNN-based and gradient-based approaches). Global model optimization-based EffQFL focuses on developing QOA for optimization and create quantum-inspired algorithms that can achieve performance improvements on specific tasks, like QAOA and VQE. On the other hand, local model optimization-based EffQFL focuses on developing exploring parameterized QNN that can adapt to the specific problem at hand, and developing hardware-aware circuit designs that consider the limitations of current QC devices, such as noise and connectivity constraints. Client selection-based EffQFL focuses on QOA that are resource-efficient, enabling the practical use of QC resources. The potential applications of EffQFL include high-dimensional data analysis and dimensionality reduction, image and signal processing, natural language processing (NLP), drug discovery and materials science, *etc*.

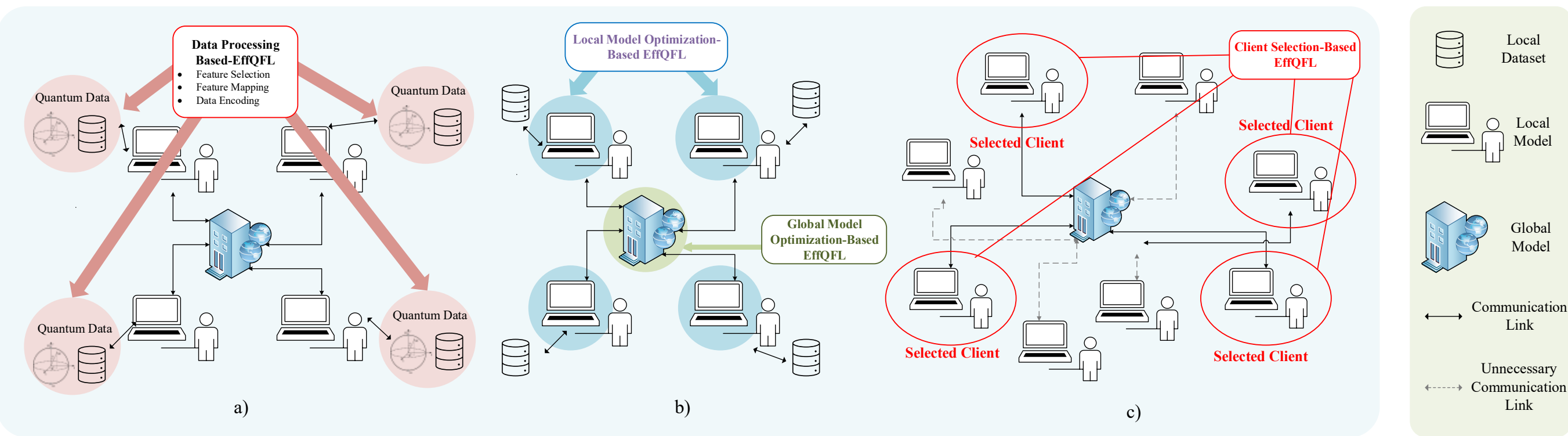

Fig. 5. Illustrations of EffQFL approaches: a) data processing-based EffQFL; b) model optimization-based EffQFL; c) client selection-based EffQFL.

In summary, EffQFL can improve the efficiency and effectiveness of feature engineering using QC. Besides, it can leverage quantum properties to create complex feature representations and speed up learning processes. Moreover, it develops resource-efficient quantum circuits and optimization algorithms and foster the transition from classical FL to QFL.

### *B. SecQFL*

SecQFL focuses on developing QFL methods that are secure and robust against adversarial attacks, quantum attacks, and unauthorized access, *etc*. One potential advantage of QFL in privacy preservation is the inherent security features of QC, such as the no-cloning theorem and the uncertainty principle, which can provide a higher level of security against certain types of cyber-attacks. Furthermore, quantum algorithms can potentially process encrypted data without decrypting it, offering a new realm of privacy-preserving computations. The primary objective is to ensure data privacy and model security while using QSM and QNN for FL process. SecQFL can mainly address the following issues:

- Data Privacy: Ensuring data privacy is essential, especially in sensitive applications like healthcare and finance.
- Model Security: Protecting FL models from unauthorized access is crucial during communications.
- Robustness: FL approaches must be resilient to noise, hardware errors, and attacks. Such SecQFL includes methods, such as QEC, that maintain reliable performance in the presence of such imperfections.

SecQFL for this category are classified into data privacy-based, model security-based, and robustness-based approaches. Data privacy-based SecQFL focuses on developing quantum privacy-preserving techniques that enable quantum feature learning while preserving data privacy, such as quantum HE (QHE), quantum DP (QDP), and quantum SMPC (QSMPC), that enable learning features from data without revealing sensitive information. Model security-based SecQFL is based on quantum cryptography techniques and utilizes quantum properties to develop secure communication protocols, such as QKD and QSDC, to protect data during transmission and storage. Robustness-based SecQFL aims to be resistant to quantum attacks, adversarial attacks, hardware imperfections, and noise by developing QSM and quantum-resistant ML algorithms, such as QEC, post-quantum cryptography, lattice-based cryptography to secure models against adversaries with QC capabilities, that can withstand attacks from adversaries with access to QC resources. The potential applications of SecQFL include secure financial transactions and fraud detection, privacy-preserving medical data analysis, cybersecurity and intrusion detection, *etc*.

In summary, SecQFL can ensure the security, privacy, and robustness of QFL methods against potential threats. Besides, it enables the adoption of quantum learning in sensitive applications where data privacy and model security are crucial. Moreover, it protects data and models from adversaries with access to QC resources and develops privacy-preserving techniques for collaborative and distributed learning scenarios.

## IV. EffQFL Category

In this section, we investigate EffQFL approaches. The principal setup and configurations for these techniques are illustrated in Fig. 5. Based on our proposed taxonomy, EffQFL approaches are divided into data processing-based, model optimization-based, and client selection-based approaches, as detailed below. Each category aims to improve the efficiency and effectiveness of QFL by leveraging quantum mechanics principles, QNN and QOA capabilities.

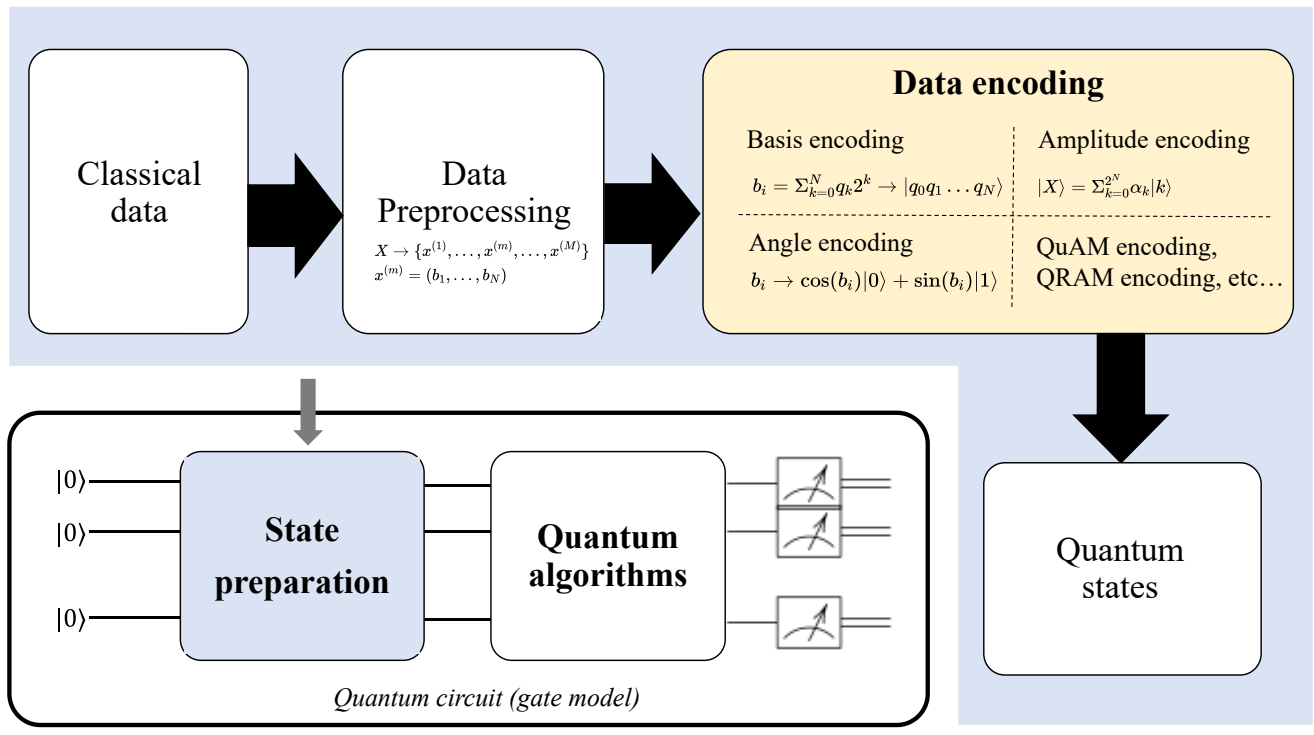

Fig. 6. Quantum encoding. (QuAM: Quantum Associative Memory; QRAM: Quantum Random Access Memory)

### *A. Data Processing-based EffQFL*

Data processing-based EffQFL approaches indeed emphasize the development of quantum encoding (shown in Fig. 6) to classical feature engineering techniques. By leveraging QC properties, these methods aim to improve the efficiency and performance of quantum feature engineering methods, mitigate overfitting, and facilitate better visualization and understanding of the data. The primary focus areas within data processing-based EffQFL include:

1) *Quantum Data Encoding:* This approach focuses on efficiently transforming classical data into quantum states by investigating different encoding schemes. The goal is to minimize the resources required for encoding and processing the data in quantum systems. Researchers investigate methods for efficiently encoding different types of data (continuous, discrete, or categorical) into quantum states while preserving their intrinsic relationships and characteristics and explore the trade-offs between the complexity of the quantum circuits required for encoding and the fidelity of the data representation. This area is analogous to classical feature extraction and preprocessing methods in feature engineering.

From information-theoretic terms, in classical communication it is impossible to transfer a continuous variable using a finite number of classical bits, without distortion, while a quantum state can be compressed with preserved entanglement with the environment using a finite number of qubits. This is since the fundamental bound is given by the Shannon entropy in classical communications, which is infinite for continuous variables. While in quantum communications, the limit is given by the von Neumann entropy which can be finite even for a continuous quantum state.

Ref. [77] proposes a statistical QFL approach to optimize non-orthogonal multiple access (NOMA) power allocation in wireless networks while maintaining user data privacy and distributing computational load. Such statistical QFL can transmit statistical information to the cloud by quantum data encoding without requiring other edges to perform NN inferences. The work in [78] use quantum data encoding method to generate quantum data on decentralized data by VQC. In [79], an input speech is up-streamed to a QC server for decentralized Mel-spectrogram feature extraction, and the corresponding convolutional features are encoded using VQC. The paper in [80] uses VQC for quantum encoding in NLP area. Ref. [81] generates the first quantum federated dataset, with a hierarchical data format, consists of excitations of quantum cluster states for large-scale and distributed quantum networks. The review in [25] also uses VQC for quantum data encoding.

2) *Quantum Feature Mapping*: Quantum feature mapping involves creating high-dimensional, non-linear representations of classical data by exploiting quantum properties. These feature maps provide a quantum version of classical feature transformation techniques, such as kernel methods. Research in this area aims to design expressive and efficient QFM that can enhance the performance of QNN models, including exploring non-linear quantum feature mapping that can generate expressive and complex feature representations, improving the resource-efficient capability of quantum models to capture the underlying structure of the data, developing techniques to select or construct optimal quantum feature mapping for specific tasks, balancing the expressiveness of the feature map and the resource requirements of the quantum circuits, and investigating the robustness of quantum feature mapping against noise and potential hardware imperfect countermeasures to improve their stability and resilience in the presence of hardware imperfections.

Quantum split learning method integrates QCNN into FL [82], which utilizes cross-channel pooling to fully leverage the unique nature of quantum state tomography made by QCNN. Quantum split learning can achieve higher accuracy than basic QFL and show many advantages in faster convergence, communication cost, and even privacy.

3) *Quantum Feature Selection and Dimensionality Reduction*: Quantum feature selection is an important aspect of data processing-based EffQFL that focuses on identifying the most relevant and informative features for a given task by strategically selecting a subset of important features. Quantum dimensionality reduction is the process of reducing the dimensionality of a dataset while preserving its essential characteristics in a QC environment.

Quantum feature selection with QFL aims to train QFL models to extract feature importance information from these models. For example, by analyzing the weights or parameters of a trained QNN, one can determine which features contribute more to the model's predictions and select them accordingly. Quantum feature selection using QOA, such as quantum variational feature selection [76], can be employed to search for the optimal subset of features that maximize a given evaluation criterion (e.g., classification accuracy, information gain). By incorporating a sparsity-inducing penalty term, quantum variational feature selection encourages the selection of a smaller subset of features while maintaining high predictive performance. Quantum principal component analysis involves finding the principal components (eigenvectors) corresponding to the largest eigenvalues of the data covariance matrix. Harrow, Hassidim, and Lloyd (HHL) algorithm [83] can be employed to compute these eigenvectors and eigenvalues, potentially speeding up the process compared to classical methods. Quantum isometric embedding aims to preserve the distances between data points in a lower-dimensional space, which leverages VQE to find an isometric embedding of the data into a lower-dimensional quantum state space. The algorithm optimizes a cost function that measures the preservation of pairwise distances between data points. Quantum locally linear embedding aims to find a lower-dimensional representation of the data while preserving the local geometry of the dataset. The local relationships between data points are preserved by reconstructing each data point as a linear combination of its neighbors in the lower-dimensional space.

### *B. Model Optimization-based EffQFL*

Model optimization-based EffQFL focuses on improving the efficiency and effectiveness of model training and aggregation through QOA. The category is divided into global and local model optimization approaches:

◆ *Global Model Optimization-based EffQFL*

Global model optimization-based EffQFL aims to optimize the overall learning process by developing QOA and quantum-inspired techniques. These approaches can achieve performance improvements on specific tasks. There are two primary sub-categories:

1) *Objective Function-based EffQFL*: These approaches aim to directly optimize the global model's objective function using QOA that guides the FL process. Researchers develop new cost functions, loss functions, or objective functions that can be optimized using QOA to improve model performance. Techniques include QAOA, VQE, and quantum metropolis sampling, which can accelerate convergence and improve model performance. QAOA can be adapted to optimize the global model's objective function by minimizing the loss function across multiple clients. Researchers can map the FL loss function to a suitable Hamiltonian and apply VQE to find the minimum energy state, which corresponds to the optimal model parameters. Quantum metropolis sampling is based on the Metropolis-Hastings algorithm, a classical Markov chain Monte Carlo method. The quantum version allows more efficient sampling of the loss function landscape in the FL setting, accelerating the optimization process. In an FL scenario for Bayesian inference, quantum metropolis sampling can be employed to efficiently explore the posterior distribution of model parameters, resulting in faster convergence and improved model performance.

To cope with various channel conditions, entangled Slimmable QFL (ESQFL) in [84] applies superposition coding across multiple depths and optimized the superposition coding power allocation by deriving and minimizing the convergence bound. This work considered the failure under extremely poor channels, importance of successful reception, NISQ limitation, and some important metrics.

2) *Model Aggregation-based EffQFL*: This subcategory focuses on using quantum version of aggregated techniques for efficient model aggregation, such as quantum FedAvg and quantum FedSGD, such quantum-inspired model aggregation techniques that take advantage of quantum parallelism to reduce the aggregation time and computational resources required for model aggregation in FL. Such approaches can be employed to perform parallel model averaging, significantly accelerating the aggregation process. For instance, in a FL setting involving multiple clients with large-scale data, quantum-inspired model aggregation can be used to simultaneously compute the weighted average of local model updates from all clients, thereby reducing the overall aggregation time.

Ref. [85] explores the non-IID issue in QFL and proposes quantum federated inference (QFedInf) method for decomposing a global quantum channel into local channels trained by each client with the help of local density estimators. This method enables to achieve one-shot communication complexity for QFL on non-IID data. Ref. [86] proposes QuanFedPS algorithm for global update based on the update unitarizes that are uploaded by each node. And this work verifies the Lemma that the order of applying update unitarizes almost does not matter and the update unitarizes almost surely exhibit the multiplicative identity property. The work in [87] applies the multiparty quantum summation by secure inner product estimation or secure weighted gradient summation (in analogy with incremental learning).

◆ *Local Model Optimization-based EffQFL*:

Local model optimization-based EffQFL concentrates on optimizing specific aspects of the QFL process at the client level. The category is divided into two main subcategories:

1) *Parameterized QNN-based EffQFL*: These methods involve designing and training parameterized QNN models, to adapt to specific problems. Researchers explore QNN for training these models efficiently and develop techniques for incorporating data processing-based approaches to improve local model performance.

The paper in [88] proposes a two-tier QFL architecture for wireless by using QNN operations in the access points and the cloud, making it an efficient solution for future wireless communications. The work in [78] uses variational quantum tensor networks and finds approximate optima in the parameter landscape. In [79], a new approach to automatic speech recognition using a QCNN is proposed with a decentralized architecture for VFL. The paper in [80] proposes a hybrid classical-quantum NN model for text classification that combines pre-trained BERT models with quantum advantages. The proposed model includes a random quantum temporal convolution (QTC) learning framework that replaces some layers in the BERT-based decoder, resulting in competitive performance on intent classification tasks. Ref. [89] uses quantum spiking NN for FL framework, called Slimmable QFL (SlimQFL). By this way, the pole and angle parameters of local quantum spiking NN models can be separately trained and dynamically communicated. Therefore, SlimQFL outperforms other QFL methods under poor communication channel conditions. The paper in [84] proposes ESQFL using a depth-adjustable architecture of Entangled Slimmable QNN (ESQNN). The review in [25] provides a framework and process for federated training on hybrid quantum-classical TL. The work in [90] is first to integrate QLSTM with temporal data for performing the task of function approximation. The work in [91] constructs the QML model to adapt the qubit connectivity of these selected subsets to guarantee the minimum number of gate insertion across all nodes. To mitigate the impact of the risks of the leaking local model information and decreasing global model efficiency, a quantum fuzzy NN is used at the local computing nodes in QFL framework [92].

2) *Gradient-based EffQFL*: This subcategory focuses on designing hardware-aware quantum circuits for gradient computation and optimization in FL. Techniques include quantum circuit cutting, which decomposes large quantum circuits into smaller, manageable pieces that can be executed on near-term quantum devices with limited qubits and noisy operations. For example, in a FL scenario for training deep QNN, clients can apply quantum circuit cutting to decompose the complex gradient computation circuits, allowing them to run on NISQ devices. Other approaches include quantum gradient descent (QGD) and quantum natural gradient (QND), which leverage QC capabilities to calculate gradients and optimize local models more efficiently than classical methods. QGD is a quantum analog of classical gradient descent algorithms, using quantum operations to compute and update model parameters more efficiently. This method can be applied to local model optimization in FL tasks, potentially accelerating convergence. In a FL setting for linear regression, clients can use QGD to efficiently update their local models' parameters, achieving faster convergence compared to classical gradient descent methods. QNG is an extension of QGD that takes into account the geometry of the quantum parameter space, leading to more efficient and robust optimization. By considering the non-Euclidean structure of the quantum parameter space, QNG can provide faster convergence and better performance in local model optimization for FL. For example, in a FL scenario for training VQC, clients can use QNG to optimize their local models by considering the unique geometry of the quantum parameter space, resulting in improved optimization and generalization.

The work in [78] uses quantum SGD for local model to avoid the analysis gradient too costly. The preprint in [93], [94] introduce federated quantum natural gradient descent (QNGD) for QFL. The FQNGD algorithm is applied in a QFL framework consisting of the VQC-based QNN. These papers show that federated QNGD can significantly reduce the communication cost among local quantum devices. Empirical studies on handwritten digit classification demonstrate the performance advantages of FQNGD over other SGD methods in theory. The work in [95] deploys QGD for private single-party gradient estimation, which can let clients of QFL architecture gain the classical gradients in each iteration. By this way, time complexity can be alleviated, achieving exponential acceleration in dataset scale and quadratic speedup in data dimensionality over the classical counterpart. Parameter shift rule, one of the most popular quantum gradient calculators, is applied to train the model. Then, the ESQNN in [84] is trained accordingly using the QGD. QuantumFed proposed in [86] enables multiple quantum nodes with node selection strategies, QuanFedNode algorithm, to collaborate using local quantum data to train a global QNN model. The framework leverages the power of QC to overcome computational power limitations in classical ML.

### C. *Client Selection-based EffQFL*

These methods focus on designing QOA that are resource-efficient, enabling the practical use of QC resources. The goal is to select the most relevant clients for FL tasks, minimizing the required communication and computation resources while maintaining high model performance. Such methods can enhance the efficiency, fairness, and adaptability of client selection, accounting for dynamic changes in the clients' data distributions, model quality, and resource availability.

1) *Quantum Clustering-based EffQFL*: These methods group clients based on their data distribution or model parameters, selecting representative clients from each cluster to participate in FL tasks, reducing communication overhead.

The work in [78] use quantum circuit cutting to approximate quantum optimization, aiming to address Ising model in FL. During this training process, QFL can collaborate with multiple clients to effectively solve the Max-Cut problem. To solve the challenges of quantum noise, limited device capacity, and high communication costs, the work in [96] designs a clustered QFL method for noisy quantum devices. This method can improve convergence through distributed task execution by circuit cutting and noise-aware device selection.

2) *Quantum Fairness and Adaptive-based EffQFL*: Such methods can be achieved by employing quantum algorithms, such as Grover's search, to efficiently sample clients based on specific criteria, like data diversity or model performance, ensuring the selected clients contribute valuable information to the global model, ensuring fairness and adaptive during the client selection process. In a FL setting with clients having diverse data contributions and model qualities, fairness-aware client selection methods can balance the participation of clients, preventing the dominance of a few clients and promoting the inclusion of underrepresented clients. Based on the changing conditions in FL systems where clients have varying data contributions and resource constraints over time, adaptive client selection methods can help maintain the system's overall performance and convergence by selecting clients that are most suitable for the current conditions.

## V. SecQFL Category

In this section, we discuss the SecQFL approaches. The principal setup and configurations for these techniques are illustrated in Fig. 7. Based on our proposed taxonomy, SecQFL approaches are divided into data privacy-based, model security-based, robustness-based approaches, as detailed below. Each category aims to improve the privacy and security of QFL by leveraging quantum mechanics principles, QSM and QML capabilities.

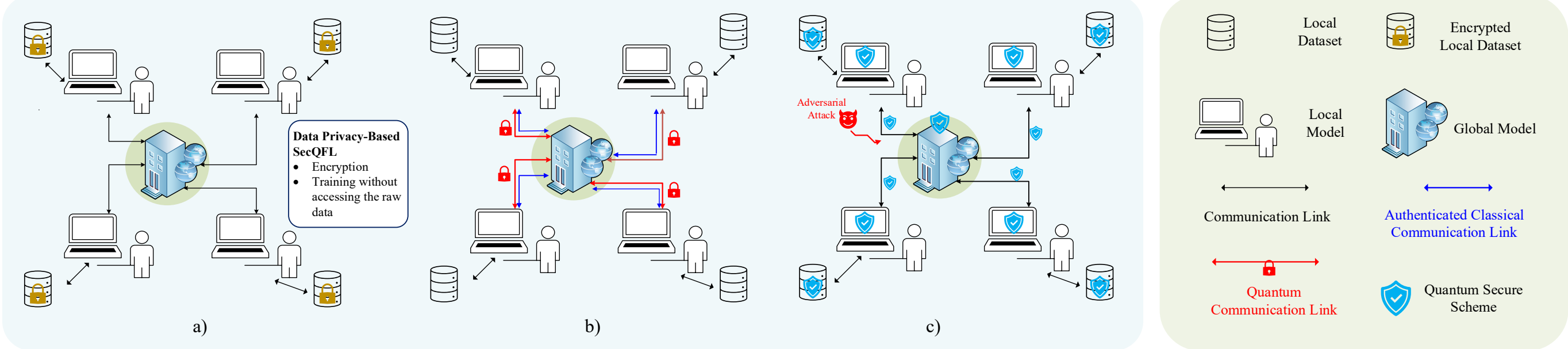

Fig. 7. Illustrations of SecQFL approaches: a) data privacy-based SecQFL; b) model security-based SecQFL; c) robustness-based SecQFL.

### A. Data Privacy-based SecQFL

These approaches focus on developing quantum privacy-preserving techniques that protect data privacy, such as QHE, QDP, and QSMPC, which enable learning features from data without revealing sensitive information.

QHE allows computations to be performed on encrypted data without the need for decryption. In FL, QHE can be used to ensure the privacy of sensitive data while still enabling clients and the server to collaboratively train a global model. Quantum fully HE (QFHE) is an advanced form of QHE that supports arbitrary computations on encrypted data. In an FL setting, clients can use QFHE to encrypt their sensitive data before sending it to the server. The server can then perform quantum computations, such as model updates, directly on the encrypted data without ever accessing the raw data. This process preserves the privacy of client data while still allowing for effective model training. QDP is a privacy framework that adds controlled quantum noise to quantum data or quantum computations to ensure the privacy of individual data points while still allowing for accurate learning from the dataset. Quantum local DP is a variant of QDP where noise is added locally by each client before sending their data or model updates to the server. This approach further enhances privacy as the server never has access to the original data or model updates. The work in [97] adds the quantum noise to the QFL model and also satisfies QDP. QSMPC allows multiple parties to jointly compute a function over their inputs while keeping the inputs private. In FL, QSMPC can be used to enable secure collaboration among clients while protecting sensitive data. Quantum oblivious transfer is a fundamental primitive in QSMPC that allows a sender to transmit one of several pieces of information to a receiver without knowing which piece was chosen. In FL, quantum oblivious transfer can be used as a building block for more complex QSMPC that protects the privacy of client data during the FL process. The work in [78] deploys QSMPC that utilizes the characteristics of quantum mechanics to address the challenges of securely calculating federated gradients.

### B. Model Security-based SecQFL

These approaches are based on quantum cryptography techniques and utilize quantum properties to develop secure communication protocols, such as QKD and QSDC, to safeguard model updates during transmission and storage. QKD protocols, such as BB84 and E91, can be used to securely distribute encryption keys between clients and a server in FL systems. By utilizing the unique properties of quantum information, QKD can detect and prevent eavesdropping attempts and ensure the security of model updates during the learning process. QSDC protocols can be employed to transmit model updates and other sensitive information securely, without the need for encryption keys. By using quantum states and properties, QSDC protocols can transmit model updates and other sensitive information securely, providing an additional layer of security in FL systems. Blind quantum computing (BQC) is a technique that enables a client to perform computations on a remote quantum server without revealing the input, output, or the computation itself. In FL systems, BQC can be used to ensure the privacy and security of the model updates, preventing unauthorized access or manipulation.

The work in [78] server uses the BQC method during server aggregation, which avoid third-party theft for server and ensure data privacy. Ref. [98] presents a quantum protocol for FL by utilizing BQC, which introduces a protocol for private single-party delegated training with VQC and extends it to multi-party distributed learning with DP. Ref. [99] proposes a new quantum secure aggregation scheme based on quantum secret sharing protocol for FL that ensures highly secure and efficient aggregation of local model parameters. The scheme uses qubits to represent model parameters, which protects private model parameters from being disclosed to semi-honest attackers. Ref. [100] applies the QSDC protocol for distributed QML that can classify two-dimensional vectors into different clusters using a remote small-scale photon quantum computation processor. Such protocol is secure and can detect any eavesdropper who attempts to intercept and disturb the learning process. The work in [101] proposes a hierarchical architecture for QKD in FL systems to optimize resource allocation and routing policies, and the work in [102] proposes a hierarchical architecture for quantum-secured federated edge learning systems with ideal security based on the QKD to facilitate public key and model encryption against eavesdropping attacks. The authors suggest a stochastic resource allocation model for efficient QKD to encrypt federated edge learning keys and models. The authors formulate and solve a stochastic programming model to achieve the optimal solution for resource allocation under uncertainties of secret-key rate requirements and weather conditions. Ref. [103] discusses the challenges faced by existing ML approaches for urban computing and proposes a hybrid FL architecture to address these challenges, which uses FL to process vast amounts of data generated from edge devices while ensuring the privacy and security of individual users' data by QKD.

### *C. Robustness-based SecQFL*

These approaches aim to make QFL models resistant to quantum attacks, adversarial attacks, hardware imperfections, and noise by developing QSM, quantum-resistant ML algorithms, and quantum-resistant cryptographic techniques. These approaches focus on developing QSM, quantum-resistant ML algorithms, and quantum-resistant cryptographic techniques that can withstand such challenges.

QEC protects quantum models from the impact of noise, hardware errors, and other imperfections in QC devices. Quantum-resistant ML algorithms, such as fault-tolerant QC and robust training methods, can make quantum models resistant to adversarial attacks and cyber-attacks. In an FL setting, QEC and quantum-resistant ML algorithms can be used to ensure that quantum model updates remain accurate and reliable. Quantum-resistant cryptography techniques can withstand quantum attacks, incorporating post-quantum cryptography and lattice-based cryptography to secure models against adversaries with QC capabilities. These algorithms aim to ensure the long-term security of FL models in the face of rapidly advancing QC technology. The work in [87] transforms the secure model aggregation task into a correlation estimation problem and generalizes the recently developed blind quantum bipartite correlator algorithm into multi-party scenarios by applying efficient redundant encoding.

In [79] and [80], random quantum circuit methods are applied in which the circuit design is randomly generated per QCNN model and QTC model for parameter protection. The proposed method in [84] controls the level of entanglement, mitigates inter-depth interference, and copes with various channel conditions by proposed fidelity-inspired regularization. Ref. [104] proposes a blockchain-based FL framework based on Lattice-based cryptography for mobile edge computing, which aims to use post-quantum cryptography to ensure that the identities of both parties are verified and quantum security. Lattice-based cryptography is safe because it is impossible to break with sufficient force in a reasonable amount of time, not even with quantum computers. The paper in [105] discusses the challenges of Byzantine attacks in QFL and proposes solutions to protect against them. The authors compare the differences between classic distributed learning and QFL, extend the Byzantine problems for the QuantumFed method [86], and theoretically define this problem. They modify the previously proposed four types of Byzantine tolerant algorithms to the quantum version and conduct simulated experiments to show a similar performance of the quantum version with the classic version. The paper in [106] proposes an optimized QFL (OQFL) framework to enhance the security in transportation systems area against adversarial attacks. The authors use a quantum-behaved particle swarm optimization technique to update the hyperparameters of FL, including the learning rate, local and global epochs. The proposed technique is utilized within a cyber defense framework to defend against adversarial attacks. The work in [97] with additional quantum noise to the QFL model not only addresses privacy leakage, but also enhances its robustness against adversarial attacks.

## VI. Implementation

To perform benchmarking effectively, it is essential to have potential platforms, various application domains, use cases, and evaluation metrics that allow for fair comparison QFL research field. In practice, QFL research field is often carried out using both real-world and synthetic datasets. In this section, we review and discuss these aspects in existing QFL literature. These benchmarking results help researchers and practitioners identify the most suitable QFL methods for specific use cases, drive further research, and improve the overall state of the field.

### *A. Platforms for QFL*

The QFL platforms are specifically designed to integrate QC techniques into FL systems or provide tools for building QFL solutions. By using these platforms, researchers and practitioners can explore the potential benefits and challenges of QFL, enabling the development of more secure, efficient, and accurate learning processes in QFL environments.

- **Qiskit**[1] [107] is an open-source QC software development kit (SDK) developed by IBM. It provides a high-level interface for designing, simulating, and running quantum circuits on quantum hardware or simulators. Qiskit consists of multiple components, such as Terra for circuit design and optimization, Aer for quantum circuit simulation, Ignis for noise modeling and error mitigation, and Aqua for quantum algorithms and applications. It is suitable for developing QFL algorithms.

- **Cirq**[2] [108] is an open-source QC software framework developed by Google. It is designed to facilitate the development, simulation, and execution of quantum circuits on various quantum processors. Cirq is part of Google's broader effort to advance QC research and applications.

- **Strawberry Fields**[3] [109] provides tools and interfaces for programming and simulating quantum circuits using continuous-variable quantum computing, a paradigm that utilizes quantum systems with an infinite-dimensional Hilbert space, such as the quantum harmonic oscillator. In

1-Webiste: https://qiskit.org/
2-Website: https://quantumai.google/cirq
3-Website: https://strawberryfields.ai/
4-Webiste: http://qutip.org/
5-Webiste: https://pennylane.ai/
6-Webiste: https://www.tensorflow.org/quantum
7-Website: https://azure.microsoft.com/en-us/products/quantum/

TABLE III
SUMMARY FOR QFL APPLICATIONS

| Application | Use Cases | References | FL Clients | FL Aggregator | Dataset |
|---|---|---|---|---|---|
| wireless communications | NOMA power allocation | [77], [88] | radio access network | core network | — |
| | | | IOT devices | cloud server | |
| | quantum sensor | [81] | data centers | 5G Infrastructure | proposed quantum dataset |
| | quantum Internet, resource allocation | [101], [102] | IOT device, edge device | cloud server | USNET topology |
| | smart manufacturing, | [113] | IOT devices | cloud server | — |
| | mobile edge computing, blockchain | [104] | data owners | server | MNIST |
| signal processing | automatic speech recognition | [79] | data owners | server | Google Speech Command V1 |
| NLP | text classification with BERTS | [80] | data centers | data centers | Snips, ATIS |
| transportation | autonomous vehicles | [106] | data centers | cloud server | MNIST |
| healthcare | clinic, digital health | [115], [116] | clinic | data centers | clinic data |
| | medical image classification | [117] | individual hospitals | aggregation servers | Non-alcoholic fatty liver disease |
| urban computing and smart cities | city data management, smart energy prediction, power management, cooperative systems | [103], [24] | data centers, smart meters, IOT devices, electric providers | cloud server, data centers | CIFAR, ImageNet |

this context, quantum information is encoded in continuous variables, such as the quadrature of an electromagnetic field, rather than discrete quantum bits (qubits).

- **QuTiP**[4] [110], which stands for Quantum Toolbox in Python, is an open-source Python library for simulating quantum systems. It is designed for simulating the dynamics of closed and open quantum systems, and it provides a wide range of tools to study quantum mechanics, quantum information theory, and quantum optics. QuTiP offers various numerical solvers for solving Schrödinger equations, master equations, and other quantum system dynamics, as well as tools for visualizing and analyzing the results.

- **Pennylane**[5] [111] is an open-source library developed by Xanadu that enables QML, QOA, and chemistry using QC hardware. It provides a unified interface for QC devices from various providers, including IBM, Google, Rigetti, and Xanadu's own photonic QC devices. Pennylane can be used to integrate QML models into FL systems, and it supports a plugin system that allows developers to create custom plugins for other QC frameworks.

- **TensorFlow Quantum**[6] (TFQ) [112] is an open-source library developed by Google. It integrates QC techniques with TensorFlow. TFQ enables the development of hybrid quantum-classical models and can be used to develop QFL algorithms. It provides a seamless interface between TensorFlow and Cirq [108], allowing users to utilize Cirq's quantum circuit capabilities within TensorFlow's ML ecosystem.

- **Microsoft Quantum Development Kit**[7] is a set of tools, libraries, and resources provided by Microsoft for quantum computing development. It includes programming languages Q#, simulators, and other components to support quantum programming and research. Q# is designed to work seamlessly with the classical .NET languages, such as C# and F#, providing a quantum development environment within the broader context of classical programming.

### *B. Applications*

In summary, QFL has shown promising results in various applications and domains, as shown in Table III, including wireless communications, NLP, urban computing and smart cities, transportation, and healthcare. There are also several benchmark datasets for corresponding QFL applications.

In wireless communications area, QFL is used to improve data processing and model optimization in wireless networks, as well as to address user data privacy concerns and distribute computational load [77], [81], [88], [101]- [104], [113]. For NLP with FL setting [114], QFL can enhance the performance and capabilities of NLP models by combining pre-trained BERT models with quantum advantages for text classification [80], which is the first mentioned QFL work in NLP area. Besides, QFL is applied in a decentralized architecture for automatic speech recognition by QCNN, addressing privacy concerns [79]. In the healthcare area, QFL has been utilized for non-identical and independently distributed clinic data [115], which is the first QFL work in healthcare area. This approach eliminates the need for data exchange, thus maintaining privacy in sensitive healthcare data. The work in [116] proposes a digital twin-assisted QFL algorithm, which can generate digital twin for patients with specific diseases for smart healthcare. In transportation systems, QFL can enhance security in transportation systems against adversarial attacks through OQFL framework [106], which is the first QFL work in transportation system area. Besides, the work in [117] solves the medical image classification problem for non-alcoholic fatty liver disease. For urban computing and smart cities, QFL has been integrated with digital twins for smart city applications [24]. The concept of "quantum digital twins" is introduced to model complex systems using QFL and improve their accuracy and efficiency. Furthermore, a hybrid FL architecture is proposed that uses QFL to process vast amounts of data generated from edge devices while ensuring privacy and security [103].

TABLE IV
EVALUATION METRICS AND PLATFORMS ADOPTED BY QFL RESEARCH

| Methods | | | **Model Performance** | | **System Performance** | | | **Trustworthy AI Performance** | | **Platforms** |
|---|---|---|---|---|---|---|---|---|---|---|
| | | | Accuracy | Convergence | Efficiency | System Scalability | Resource Utilization | Robustness | Interpretability | |
| **EffQFL** | data processing-based | | [25], [78]-[82] | [25], [78] | [77], [81] | [81], [82] | [79], [82] | - | [79] | PennyLane, Qiskit, TFQ, Torchquantum |
| | model optimization-based | global | [84]-[86] | [84], [85] | - | - | - | [84], [86] | - | Tensorcircuit, QuTip |
| | | local | [25], [78]-[80], [84], [86], [89]-[94] | [25], [78], [84], [93]-[95] | [89] | [89] | [79] | [84], [86] | [79] | PennyLane, TFQ, Qiskit, QuTip |
| | client selection-based | | [78] | [78] | - | - | - | - | - | TFQ |
| **SecQFL** | data privacy-based | | [78] | [78] | - | - | - | - | - | TFQ |
| | model security-based | | [78], [98], [99] | [78], [102] | [103] | - | [101]-[103] | [98], [99] | - | TFQ, GAMS/CPLEX |
| | robustness-based | | [79], [80], [84], [104], [106] | [84], [104] | [104] | [104] | [79] | [84], [104]-[106] | [79] | PennyLane, Qiskit, QuTip |

These applications demonstrate the potential of QFL to address various challenges and deliver improvements in multiple domains, from enhancing security and privacy to increasing efficiency and performance. As QC technology continues to advance, we can expect more breakthroughs and applications of QFL in a wide range of industries.

### C. *Evaluation Metrics*

We divide the evaluation metrics for QFL research into three categories: 1) model performance, 2) system performance, and 3) trustworthy AI-related, as shown in Table IV.

Model performance-related metrics in QFL research focus on evaluating the accuracy and convergence of the models. These metrics help assess the effectiveness of the learning process and its ability to generalize well on unseen data. Accuracy metric for QFL [25], [78]-[86], [89]-[94], [98], [99], [104], [106] consists of average training, validation and testing accuracy, and global model and local model accuracy. Convergence measures how well the model is learning and reaching an optimal solution. It is essential to ensure that the model training process is stable and converging to a good solution, including parametric landscape [78], loss [25], [78], [84], [85], cost function value [93], [94], objective function value [78], [95], Taylor loss [102], convergence time [104].

System performance-related metrics in QFL focus on evaluating aspects such as communication and computational efficiency, system scalability, and resource utilization. These metrics help assess the overall performance of the QFL system in terms of resource usage and reliability. Efficiency metric consists of communication efficiency (evaluating the effectiveness of the communication process between clients and the server) and computational efficiency (measuring the computation time), including throughput [77], number of epochs [81], [89], CPU execution time [103], and delay [104]. System scalability assesses the ability of QFL system to accommodate the adaptive and dynamic clients and maintain its performance [81], [82], [89], [104]. Resource utilization metrics focus on assessing the consumption of resources like computation, communication, memory, and energy during the learning process, including parameters memory and memory usage [79], [103], wavelength utilization [101], communication cost [82], and deployment cost [101], [102].

Trustworthy AI metrics have not been widely implemented for evaluating QFL approaches. However, there are some emerging works that take these metrics into account [79], [84], [86], [98], [99], [104]-[106]. In one study [79], interpretable neural saliency is employed to assess the interpretability performance of the proposed method. Fidelity index in [84], [86], [98], [99], [105], [106] is applied for robustness performance. The work in [104] uses attack detection rate and cumulative reward to measure the robustness against potential threats.

## VII. CHALLENGES, OPEN OPPORTUNITIES AND PROMISING FUTURE RESEARCH DIRECTIONS

The field of QFL is still in its infancy, and numerous challenges remain to be addressed. These include quantum hardware limitations, noise and error mitigation, model and data heterogeneity, interoperability between quantum and classical FL, standardization and benchmarking, and ethics and legal considerations. However, as QC technology continues to advance, we can expect the field to evolve and mature rapidly. In the coming years, QFL may become an essential tool for addressing complex problems and preserving data privacy in an increasingly interconnected world. As researchers and practitioners continue to explore the potential of QFL, it will be crucial to keep an eye on new developments and breakthroughs in both QC and FL to ensure that the full potential of this promising field is realized.

### *A. Quantum Hardware Limitations and Resource Constraints*

Quantum hardware is still in its early stages of development, with limited qubits, high noise levels, and relatively short coherence times. These limitations pose challenges for implementing complex QFL algorithms, necessitating ongoing research and development to improve quantum hardware capabilities and make QFL more practical.

- *Limited Qubits*: Current quantum computers have a limited number of qubits, which restricts the size and complexity of quantum circuits that can be executed. The development of fault-tolerant quantum computers with a larger number of qubits is crucial for scaling up QFL algorithms.
- *Coherence Time*: Quantum states in a quantum computer are fragile and can only be maintained for a short period, known as the coherence time. Increasing coherence time is necessary to execute more complex QFL algorithms without significant loss of information.
- *Connectivity*: Limited qubit connectivity in current quantum hardware can impact the efficiency of quantum algorithms. Designing QFL algorithms that take into account hardware constraints, such as qubit connectivity, can improve their performance.

### *B. Noise and Error Mitigation*

Quantum systems are inherently susceptible to noise and errors due to their delicate nature. Effective error mitigation techniques are crucial for improving the accuracy and reliability of QFL algorithms. Research into novel error correction and mitigation methods, as well as the development of more robust quantum algorithms, can help address these challenges.

- *Error Correction*: Developing efficient and fault-tolerant QEC codes that can correct errors while maintaining low overheads is important for improving the reliability of QFL algorithms.
- *Error Mitigation Techniques*: Novel error mitigation techniques, such as noise extrapolation and error-aware training, can help enhance the performance of QFL algorithms in the presence of noise.

### *C. Model and Data Heterogeneity*

FL involves heterogeneous models and data distributions across participating devices, which may affect the performance of QFL algorithms. Addressing model and data heterogeneity requires the development of adaptive and scalable quantum algorithms that can handle such diversity effectively.

- *Adaptive Algorithms*: Developing QFL that can adapt to varying model architectures and data distributions is essential for addressing heterogeneity issues in FL environments.
- *Decentralized Optimization*: Investigating decentralized optimization techniques for QFL can help improve the convergence of the algorithms in heterogeneous settings.

### *D. Interoperability between QFL and Classical FL*

Seamless integration of quantum-enhanced FL with classical FL methods is essential for their widespread adoption.

- *Hybrid Algorithms*: Research into hybrid quantum-classical algorithms that can be easily integrated into existing classical FL systems can facilitate a smoother transition to quantum-enhanced FL.
- *Middleware Solutions*: Developing middleware solutions that enable seamless communication and data exchange between classical and quantum systems can help bridge the gap between the two paradigms.

### *E. Standardization and Benchmarking*

To facilitate the comparison and evaluation of different QFL algorithms, standardization and benchmarking are essential. Establishing common datasets, performance metrics, and benchmark protocols can help researchers and practitioners assess the effectiveness of various QFL techniques and promote their adoption.

- *Benchmark Datasets*: Establishing standardized datasets for QFL will enable fair comparisons between different algorithms and promote better understanding of performance.
- *Performance Metrics*: Defining suitable performance metrics for QFL, such as communication cost, convergence rate, and model accuracy, will help assess the effectiveness of QFL.
- *Benchmark Protocols*: Developing robust benchmark protocols, including hardware specifications and algorithm configurations, can ensure reliable and consistent evaluation of QFL algorithms.

### *F. Ethics and Legal Considerations*

As with any technology, trustworthiness considerations must be taken into account when developing and deploying QFL systems. Ensuring data privacy, security, and compliance with regulatory requirements is essential for the responsible use of these technologies. Additionally, researchers and practitioners must consider the potential impact of QFL on fairness, accountability, transparency, and stakeholder engagement in ML models.

- *Data Privacy*: Ensuring that QFL algorithms preserve data privacy and adhere to relevant data protection regulations is crucial for responsible technology deployment.
- *Fairness*: Employing quantum techniques to assess fairness in FL systems can help identify and mitigate potential biases, ensuring that the learning process is more equitable and representative of the diverse clients participating in the network. Developing methods to improve fairness in QFL algorithms, such as re-sampling techniques, adversarial training, and fair optimization, can help prevent discriminatory outcomes.
- *Accountability and Transparency*: Establishing guidelines and best practices for the responsible development, deployment, and monitoring of QFL systems can help ensure accountability and transparency in the field. Research into interpretability (XAI) techniques for QFL algorithms can contribute to a better understanding of their decision-making process and promote trust in these systems.

➢ *Stakeholder Engagement*: Engaging with stakeholders, including policymakers, industry partners, and end-users, can help raise awareness about the ethical and legal implications of QFL and facilitate the development of responsible guidelines and regulations.

### *G. Promising Future Research Directions of QFL*

As the field of QFL continues to evolve, these open opportunities and future directions will guide researchers and practitioners in their pursuit of novel solutions, innovative approaches, and groundbreaking applications that harness the power of QC and FL for a greater AI paradigm. We acknowledge the importance of addressing both the long-term potential and the immediate challenges in QFL. While we maintain an optimistic outlook on the future advancements and applications of QFL, we also recognize the necessity of focusing on more short-term, practical issues that are currently impeding progress in this field. These short-term challenges are given as below.

Firstly, current QC technology faces limitations in creating deep quantum models that are analogous to DNN in classical ML. Addressing the complexity and resource requirements for such models is crucial for advancing QFL. Secondly, similar to classical NNs, QNNs also face challenges in gradient calculation, which is essential for the training and optimization of these models. Developing efficient and accurate methods for gradient computation in QNNs is a pressing need. Thirdly, the vanishing gradient problem, a well-known issue in DL, is also present in QNNs. This issue becomes more pronounced as the depth of quantum circuits increases, hindering the effective training of QNNs. Fourthly, effective integration between classical and quantum computations remains a significant challenge. This includes data encoding, algorithm design, and harnessing the strengths of both quantum and classical computing paradigms to optimize computational tasks.

Additionally, our summary in Table II, highlighted by “○”, also indicates potential short-term directions for QFL development. These research areas, while currently underexplored or having less existing work, represent fertile ground for future exploration.

Next, we also provide some promising long-term future research directions for QFL as below.

➢ Personalized QFL involves applying QC techniques to FL in a manner that is individually tailored for individual users, suitable for devices such as smartphones or IoT systems. This personalization is achievable as each local model is designed to cater to the unique data of its device, while simultaneously drawing insights from the aggregate learning of all devices involved [118]. In a quantum-enhanced setting, this process could be significantly enhanced by the capabilities of QNN and QOA. Such enhancements could lead to more effective learning outcomes and improved user experiences, a critical factor in sectors like healthcare and finance where data privacy and processing efficiency are of utmost importance. The integration of QFL into these devices promises not only improved learning capabilities but also enhanced user experiences through more sophisticated data analysis and decision-making processes. While the integration of QFL into usual network edge or IoT systems is a complex and evolving endeavor, it represents a significant step forward in computational technology. The potential benefits of such integration are vast, offering more personalized, efficient, and secure data processing capabilities in a wide range of applications.

➢ Quantum federated graph NN (QFGNN) cutting refers to the application of QFL to graph neural networks, specifically for solving Max-Cut problems. The Max-Cut problem is a well-known problem in computer science and combinatorial optimization, where the goal is to divide the nodes of a graph into two groups so as to maximize the sum of the weights of the edges crossing between the groups [119]. In the QFGNN, this could have practical applications such as node selection in a FL system, judging the connectivity of a QFGNN, and reducing redundant edges while ensuring the graph structure remains connected.

➢ Quantum-inspired compression and quantization techniques can help reduce the size of model updates and the overall communication overhead in FL systems. QFL can leverage quantum data encoding to compress model updates and transmit information more compactly than classical FL methods. This can enable more efficient model updates and faster convergence. Additionally, QFL can use quantum-inspired techniques for efficient model quantization, which involves reducing the precision of model parameters to reduce the size of model updates. This can help reduce the communication overhead and improve the efficiency of FL systems without sacrificing model performance.

➢ Large-scale language models, such as GPT-3 or GPT-4, embody potent instruments capable of harnessing extensive general knowledge. Therefore, these models are excellent in discerning contextual for generating contextually relevant responses [120]. Nonetheless, these pioneering models present certain challenges that warrant meticulous attention and further exploration. A pivotal domain ripe for additional research pertains to the potential role of QFL in bolstering the scalability and efficiency of language models, particularly in the quantum natural NLP domain. This includes innovating QFL algorithms tailored for language modeling. It also involves exploring the intricate interplay between QC and other burgeoning technologies such as blockchain and edge computing. We recognize the scalability challenges faced by QC in handling massive datasets, akin to those used in state-of-the-art language models like GPT-3 and GPT-4, particularly within the constraints of the NISQ era. The fusion of these technologies could potentially drive unprecedented advancements in the realm of large-scale language modeling, paving the way for a new era of computational linguistics.

➢ Incorporating multimodal QFL into our future directions aligns with the evolving landscape of quantum computing and machine learning. The future of QFL lies in its ability to integrate and process information from multiple data modalities, including text, images, audio, and sensor data. Multimodal QFL leverages the strengths of quantum computing to handle high-dimensional data efficiently, enhancing model robustness and performance. Privacy and security remain paramount, with quantum encryption techniques ensuring the confidentiality of diverse data types. Potential applications span healthcare, smart cities, finance, and beyond, where multimodal QFL can lead to more accurate diagnostics, improved urban management, and enhanced decision-making processes. Exploring multimodal QFL represents a significant step towards realizing the full potential of QFL.

Overall, by focusing on these immediate challenges, along with exploring the long-term potential of QFL, we aim to contribute to a more comprehensive and pragmatic development of the field. This balanced approach should pave the way for more immediate applications of QFL, while also laying the groundwork for future advancements."

# VIII. Conclusions

This review aims to provide a comprehensive understanding of the principles, techniques, and emerging applications of QFL. This interdisciplinary approach aims to leverage the strengths of quantum technologies to enhance privacy, security, and efficiency in the learning process. The review includes an overview of FL, quantum algorithms, and motivations for QFL. This review is the first work for providing a taxonomy of QFL techniques and highlights existing works in the field. The review also discusses evaluation metrics, platforms, and datasets used in QFL research. Finally, this review identifies challenges and outlines future research directions and open research questions, focusing on areas like quantum hardware limitations, noise mitigation, model heterogeneity, and ethical considerations. This review serves as a guide for those interested in QFL.